\documentclass[review]{elsarticle}
\graphicspath{ {./} }
\usepackage{hyperref}
\usepackage{float}
\usepackage{verbatim} 
\usepackage{apalike}
\restylefloat{figure}
\restylefloat{table}
\usepackage{mathtools, array, booktabs}
\usepackage{tabularx}
\usepackage{pdflscape}
\usepackage{afterpage}
\usepackage{lipsum}

\usepackage{amsmath}
\usepackage{multicol}
\usepackage{caption}
\usepackage{graphicx}
\usepackage{textcomp}
\usepackage{amsfonts}

\usepackage{makecell}
\usepackage{siunitx}
\usepackage{etoolbox}
\usepackage{pdfpages}
\usepackage{wrapfig}
\usepackage{todonotes}
\usepackage{subcaption}
\usepackage{dirtytalk}
\usepackage{multirow}

\usepackage[toc,page]{appendix}
\usepackage{chngcntr}

\DeclareMathOperator*{\argmin}{arg\,min}

\usepackage[ruled,vlined]{algorithm2e}
\usepackage{algpseudocode}

\let\today\relax
\makeatletter
\def\ps@pprintTitle{%
    \let\@oddhead\@empty
    \let\@evenhead\@empty
    \def\@oddfoot{\footnotesize\itshape
         {This work was accepted for publication in Neural Network.  Email: youbanghust@gmail.com} \hfill\today}%
    \let\@evenfoot\@oddfoot
    }
\makeatother

\biboptions{authoryear}

\begin{document}\sloppy
\begin{frontmatter}










\title{Multimodal Information Bottleneck for Deep Reinforcement Learning with Multiple Sensors}

\author[label1]{Bang You}
\author[label1]{Huaping Liu}

\address[label1]{Department of Computer Science and Technology, Beijing National Research Centre for Information Science and Technology, Tsinghua University, Beijing 100084, China}

\begin{abstract}
Reinforcement learning has achieved promising results on robotic control tasks but struggles to leverage information effectively from multiple sensory modalities that differ in many characteristics. Recent works construct auxiliary losses based on reconstruction or mutual information to extract joint representations from multiple sensory inputs to improve the sample efficiency and performance of reinforcement learning algorithms. However, the representations learned by these methods could capture information irrelevant to learning a policy and may degrade the performance. We argue that compressing information in the learned joint representations about raw multimodal observations is helpful, and propose a multimodal information bottleneck model to learn task-relevant joint representations from egocentric images and proprioception. Our model compresses and retains the predictive information in multimodal observations for learning a compressed joint representation, which fuses complementary information from visual and proprioceptive feedback and meanwhile filters out task-irrelevant information in raw multimodal observations. We propose to minimize the upper bound of our multimodal information bottleneck objective for computationally tractable optimization. Experimental evaluations on several challenging locomotion tasks with egocentric images and proprioception show that our method achieves better sample efficiency and zero-shot robustness to unseen white noise than leading baselines. We also empirically demonstrate that leveraging information from egocentric images and proprioception is more helpful for learning policies on locomotion tasks than solely using one single modality.
\end{abstract}

\begin{keyword}
Reinforcement learning \sep Representation learning \sep Multimodal data \sep Multisensor fusion \sep Information bottleneck
\end{keyword}

\end{frontmatter}

\section{Introduction}
\label{introduction}



Developing reinforcement learning (RL) agents capable of combining information from multiple sensors is important for performing complex robotic control tasks~\citep{lee2020making, fazeli2019see}. For example, a locomotion robot needs to observe obstacles in front of its body via its egocentric camera and sense its own motors by using force and torque sensors.  Since the information from these modalities is complementary and concurrent, leveraging multimodal sensory inputs can help perceive the environment~\citep{liang2021multibench, yang2021learning, cong2022reinforcement} and be useful when one modality is noisy~\citep{lee2020multimodal}. Despite the comprehensible benefits of combining egocentric images and proprioceptive feedback, deep reinforcement learning agents struggle to learn a good policy from multiple sensory modalities that have different physical properties and dimensionality~\citep{hansen2022visuotactile}. Using multimodal observations for RL can even fail to match the performance of using a single modality~\citep{chen2021multi}.

Constructing an auxiliary loss to learn compact representations from raw observations along with policy learning has been proven to effectively improve the sample efficiency and performance of RL algorithms~\citep{lesort2018state}. The proposed auxiliary losses include reconstructing observations~\citep{yarats2019improving, watter2015embed}, data augmentation~\citep{yarats2021image}, future prediction~\citep{lee2020predictive, stooke2021decoupling, you2022integrating}, and contrastive learning~\cite{laskin2020curl, lee2023stacore}. However, most of these approaches only investigate the effectiveness of the proposed losses on tasks with third-person-view images, and have no guarantee of inducing an appropriate multimodal representation from egocentric images and proprioception that have different characteristics.

A commonly used technique for combining the information from images and proprioception is to concatenate proprioceptive data with representations solely extracted from images~\citep{cong2022reinforcement, yang2023neural}, since proprioceptive feedback is already compact. Some recent approaches map multimodal observations into a joint latent space and learn joint representations by minimizing a reconstruction error~\citep{hafner2023mastering, wu2023daydreamer}. Furthermore, \citet{chen2021multi} propose to learn a joint representation from multimodal observations by maximizing the mutual information between the joint representation and representations extracted from single modalities. However, representations learned by these methods have no incentive to discard irrelevant information (e.g. noise or background changes) in raw multimodal observations~\citep{lee2020predictive, fan2022dribo}, which may degrade the performance of RL agents.

We argue that compressing information in extracted joint representations about multiple sensory inputs is important, since most information in raw multimodal observations, especially in noisy ones, is unrelated to learning a policy. Performing the compression can force the latent joint representations to retain only the task-relevant information.

Hence, in this paper we propose a multimodal information bottleneck model based on the information bottleneck principle~\citep{tishby1999information}, MIB, which learns task-relevant joint representations from egocentric images and proprioception for reinforcement learning. Our MIB model aims to learn compressed joint representations, which fuse and preserve complementary information from multiple sensors, and meanwhile, filter out information in raw multimodal observations unrelated to learning a policy. To this end, our MIB model compresses the information between the latent joint representation and raw multimodal sensory inputs, while maximizing the predictive information between two successive joint representations given the corresponding action. We apply variational inference to derive an upper bound of our MIB objective for computationally feasible optimization.

We train a standard Soft Actor-Critic (SAC) agent~\citep{haarnoja2018soft} along with our MIB model for learning a continuous policy from multimodal sensory inputs. We conduct experiments on several high-dimensional locomotion tasks with egocentric images and proprioceptive information. We also evaluate the zero-shot robustness of the learned representations and policies to observation perturbations, both white noise and natural backgrounds.  The main contributions of our paper are summarized as follows:

\begin{itemize}
    \item We propose a multimodal information bottleneck model and use it to learn compressed joint representations from raw images and proprioception.
    \item We propose a variational upper bound of our MIB objective for tractable optimization.
    \item We demonstrate that integrating information from egocentric images and proprioception for learning policies improves the performance on challenging locomotion tasks compared to solely using one single modality.
    \item Empirical results show that our method achieves better sample efficiency on locomotion tasks, while being more robust to unseen white noise and natural backgrounds compared to leading baselines.
    
\end{itemize}

The remainder of the paper is organized as follows. In Section~\ref{related work} we discuss the related work. In Section~\ref{preliminaries} we present the problem statement and notations. We present our MIB model in Section~\ref{method} and experimental setup and evaluations in Section~\ref{exp}. In Section~\ref{conclusion} we summarize our work and discuss future works.

\section{Related Work}
\label{related work}

In this section, we discuss the related works to our method. We will discuss multimodal representation learning methods in RL. We also next discuss information-theoretic representation learning approaches in image-based RL. 

\subsection{Multimodal Representation Learning in RL}

Motivated by the success of multimodal fusion~\citep{liu2017multimodal, liu2018robotic, li2021research, wang2021multi} and machine learning~\citep{qiao2023robotic, kang2023manual,zhang2023efficient}, some recent works learn RL policies from multimodal sensor data with different characteristics~\citep{lee2020making, liu2023hybrid, hansen2022visuotactile, yasutomi2023visual, cho2022s2p, yang2021learning, loquercio2023learning}. Our work focuses on learning an optimal policy from egocentric images and proprioceptive information. Learning representations solely from images and then concatenating compact image representations and proprioception is an effective method to leverage complementary information from images and proprioception~\citep{cong2022reinforcement,yang2023neural}. Differing from the concatenation of image representations and proprioception, our method learns a joint multimodal representation by mapping multiple modalities into a joint latent space.  \citet{noh2022toward} extract representations from RGB images and proprioception separately and then learn a joint representation from these two extracted representations for the RL agent. They optimize the latent joint representation using only the reward signal. Instead, some approaches~\citep{hafner2023mastering, wu2023daydreamer, hafner2022deep} formulate the problem of learning multimodal joint representations as auto-encoding variational inference, where the representations are trained by reconstructing the raw observations. In order to avoid capturing task-irrelevant details in the raw observation by reconstructing, MuMMI~\citep{chen2021multi} proposes to replace the reconstruction loss with the mutual information maximization objective, which maximizes the mutual information between representations extracted from single modalities and the latent joint representation. To estimate this mutual information, MuMMI employs the InfoNCE loss with a score function that encourages representations of single modalities to be close to the joint representation. Similarly, \citet{becker2023reinforcement} integrates the reconstruction loss and the contrastive loss for learning latent joint representations from images and proprioception. In contrast to these methods based on reconstruction and mutual information, we fuse and \emph{compress} information from multiple sensors for learning joint representations. Our method minimizes the mutual information between the learned joint representation and the image and proprioceptive representations to avoid capturing much information in raw observations that is not useful for learning a policy.

\subsection{Information-Theoretic Representation Learning in RL}
Many recent approaches use information-theoretic objectives as auxiliary tasks to learn compact latent representations from images for image-based reinforcement learning~\citep{oord2018, anand2019unsupervised, you2022integrating, rakelly2021mutual}. A common objective is to maximize mutual information between representations of states. For instance, CURL~\citep{laskin2020curl} maximizes the shared information between the representations extracted from two independently augmented observations, which are generated by applying random crop data augmentation. \citet{stooke2021decoupling} propose to capture the temporal predictive information between the representation of two successive states within the same trajectory for learning state representations. The encoders that extract representations from states are decoupled from the policy learning during optimization. ~\citet{mazoure2020deep} capture the predictive information between representations of sequential states given the corresponding action for improving the concordance of the latent representations. Instead of directly maximizing the mutual information between representations, our method is more closely related to methods based on the information bottleneck principle~\citep{tishby1999information} that provide an explicit mechanism to compress task-irrelevant information in observations. ~\citet{lee2020predictive} apply the conditional entropy bottleneck objective proposed by Fischer et al.~\citep{fischer2020conditional} to preserve the mutual information between the compressed representation and the future states and rewards, while filtering out the information from the previous states and future actions that is irrelevant to the future states and rewards. DB~\citep{bai2021dynamic} learns a compressed embedding by maximizing the mutual information between the compressed embedding and the embedding of the next state, and meanwhile minimizing the mutual information between the compressed embedding and the representation of the current state and the action. ~\citet{fan2022dribo} learn latent representations in the multi-view setting by retaining the shared information between representations of sequential multi-view observations and discarding the information not shared by these observations. The multi-view observations are obtained by applying random data augmentation on the raw observations. However, these aforementioned methods based on information bottleneck learn the latent representation solely from images. The effectiveness of compression
in control tasks with multimodal observations has not been verified yet. Instead, we aim to learn a joint representation from multimodal observations by compressing and preserving the predictive information from multiple sensory inputs. We analyze the relation between performing compression on multimodal observations and the downstream performance of the RL agent.

\section{Preliminaries and Notations}
\label{preliminaries}
\subsection{Problem Statement and Notations}
We formulate the problem of learning a continuous policy from multimodal observations as an infinite-horizon Markov decision process (MDP) described by the tuple $\mathcal{M} =(\mathcal{S},\mathcal{A},P,r,\gamma)$, where $\mathcal{S}$ is the state space, and $\mathcal{A}$ is the action space, $P(s_{t+1}|s_t,a_t)$ is the transition model, $r(s,a)$ is the reward function and $\gamma \in (0,1)$ is the discount factor. At every time step, the agent observes the current state $s_t$ and selects the action $a_t$ by using its stochastic policy $\pi(a_t|s_t)$, and then obtains the reward $r(s_t,a_t)$. The goal is to optimize the policy to maximize the cumulative discounted reward. 

We focus on reinforcement learning from multiple sensors, where the state space is provided in terms of proprioception and egocentric images. The multimodal observations at time step $t$ can be denoted as ${s_t = \{o_t^m\}}_{m=p,i}$, where $o_t^p$ is the proprioception at time step $t$ and $o_t^i$ is the image observation that stacks the three most consecutive frames by following~\citep{yarats2019improving, noh2022toward}. The observation from an individual modality may be high-dimensional, noisy, and not sufficient to describe the underlying state of the environment. We assume the multimodal observation $s_t$ contains all the information necessary for decision-making. Our goal is to find a mapping that encodes the high-dimensional and multimodal observation $s_t$ into a latent joint representation $j_t$ to improve performance and sample efficiency. Along with representation learning, we aim to learn a policy that maximizes the cumulative discounted rewards
\begin{equation}
    J(\pi) = \mathbb{E}_{\pi} \left [ \sum_{t=0}^\infty \gamma ^ t r_t \bigg | a_t \sim \pi(\cdot | j_t), s_{t+1} \sim P(\cdot | s_t, a_t), s_0 \sim p_0(s_0)\right]
\end{equation}
with the discount factor $\gamma \in (0,1)$ and  the distribution of the initial multimodal state $p_0(s_0)$.

\section{Multimodal Information Bottleneck}
\label{method}
In this section, we present the proposed multimodal information bottleneck model and how to train our model along with the policy and value function.

\subsection{The Multimodal Information Bottleneck Objective}

Our multimodal information bottleneck model aims to extract compressed joint representations from proprioception and egocentric images. We denote the embeddings of the current image observation $o_t^i$ and  the embeddings of the current proprioception $o_t^p$ as $c_t^i$ and $c_t^p$, respectively. The goal of our multimodal information bottleneck model is to obtain a compressed joint representation $z_t$ from $c_t^i$ and $c_t^p$, which discords the task-irrelevant information from the raw visual and proprioceptive observations and only retains the relevant predictive information for learning a policy. 

\begin{figure*}
    \centering
    \centerline{\includegraphics{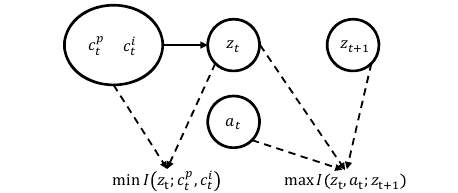}}
    \caption{Illustration of the objective of our MIB model. Our MIB objective minimizes the mutual information between the joint representation $z_t$ and the embeddings of the current image and proprioception for compression, while maximizing the predictive information $I(z_t, a_t; z_{t+1})$ for improving the latent temporal consistency.}
    \label{fig:mib objective}
\end{figure*}

Following the information bottleneck principle~\citep{tishby1999information, alemideep},  our MIB model learns the compressed joint representation $z_t$ by minimizing the mutual information between the bottleneck variable $z_t$ and the image and proprioception representations and meanwhile preserving the relevant predictive information between the current bottleneck variable and the action and the next bottleneck variable. To this end, the objective of our MIB model is to solve the optimization problem

\begin{equation}
\begin{aligned}
\min \quad \alpha I(z_t;c_t^p, c_t^i) - I(z_t, a_t; z_{t+1})
\end{aligned}
\label{eq:mib objective},
\end{equation}
where the hyperparameter $\alpha >0 $ balances compression and relevance. The mutual information $I(z_t;c_t^p, c_t^i)$ measures the dependency between the bottleneck variable and the image and proprioception representations, which has the form

\begin{equation}
\begin{aligned}
I(z_t;c_t^p, c_t^i) &= \mathbb{E}_{p(z_t,c_t^p, c_t^i)} \bigg[\log \frac{p(z_t,c_t^p, c_t^i)}{p(c_t^p, c_t^i) p(z_t)}\bigg]\\
\end{aligned},
\end{equation}
where the expectation is computed over the joint distribution $p(z_t,c_t^p, c_t^i)$. The predictive information $I(z_t, a_t; z_{t+1})$ qualifies the decreased uncertainty of $z_{t+1}$ given $z_t$ and $a_t$, which has the form
\begin{equation}
\begin{aligned}
I(z_t, a_t; z_{t+1}) &= \mathbb{E}_{p(z_t, a_t, z_{t+1})} \bigg[\log \frac{p(z_t, a_t,z_{t+1})}{p(z_t, a_t) p(z_{t+1})}\bigg].\\
\end{aligned}
\end{equation}
Intuitively, minimizing the mutual information $I(z_t;c_t^p, c_t^i)$ forces the latent variables to filter out task-irrelevant information in raw image and proprioception as much as possible, such as noise. Maximizing the predictive information $I(z_t, a_t; z_{t+1})$ encourages the bottleneck variables $z_t$ to retain sufficient information for improving temporal consistency in the latent space, which is useful for learning a policy~\citep{zhao2023simplified}. Notably, our MIB objective in Eq.~\ref{eq:mib objective} aims to jointly compress \emph{multimodal} observations,  while prior works~\citep{bai2021dynamic, lee2020predictive} in RL only compress  \emph{individual} modalities. Moreover, instead of preserving the predictive information between the bottleneck variable and the representation of the next state~\citep{bai2021dynamic}, our MIB objective captures the predictive information between representations of consecutive states given the corresponding action. This predictive information is associated with forward dynamics in the latent space, which can further encourage the latent bottleneck variable to discord information that is irrelevant to policy learning.

\subsection{Model Overview}
Unfortunately, the objective of our MIB model is intractable to directly compute. To solve this problem, we derive an upper bound of the objective (Eq.~\ref{eq:mib objective}), by obtaining an upper bound of the mutual information $I(z_t;c_t^p, c_t^i)$ and a lower bound of the predictive information $I(z_t, a_t; z_{t+1})$. Before presenting the derived upper bound of our MIB objective, we provide an overview of the proposed network architecture of our MIB model.

\begin{figure*}
    \centering
    \makebox[\textwidth]{\includegraphics[width=\textwidth]{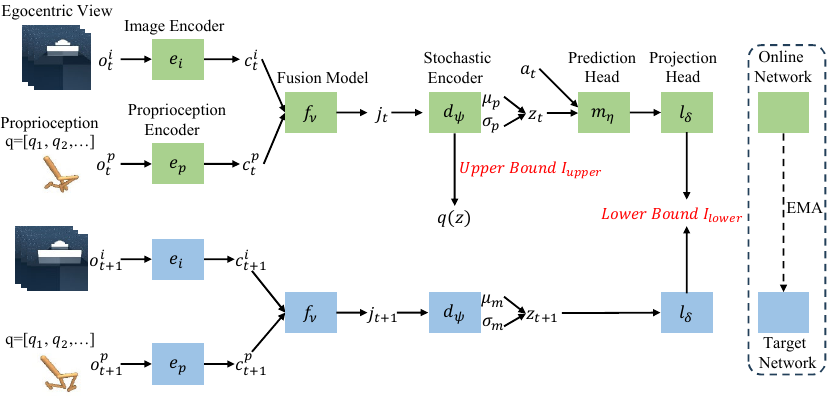}}
    \caption{The network architecture of our MIB model. We use an image encoder and a proprioception encoder to extract latent representations from the current egocentric image and proprioception, respectively. The obtained upper bound of the mutual information $I(z_t;c_t^p, c_t^i)$ is the KL divergence between the distribution of the joint representations $z_t$ given the extracted representations and the unit normal Gaussian distribution. The prediction head and the projection head are used to map the joint representations in a latent space, where the lower bound of the mutual information $I(z_t, a_t; z_{t+1})$ is computed.}
    \label{fig:network}
\end{figure*}

Our multimodal information bottleneck (MIB) model constructs four components for generating latent representations:
\begin{equation}
\begin{aligned}
\text{Image and Proprioception Encoder} &: \quad c_t^i = e_{i}(o_t^i;\theta_i), \quad c_t^p = e_{p}(o_t^p;\theta_p) ,\\
\text{Fusion Model} &: \quad j_t = f_\upsilon(c_t^i, c_t^p; \theta_\upsilon),\\
\text{Stochastic Encoder} &: \quad z_t \sim d_\psi(z_t|c_t^p, c_t^i; \theta_\psi).
\end{aligned}
\label{eq:encoding}
\end{equation}
We first use the image encoder $e_i$ with parameter $\theta_i$ to map the current image observation $o_t^i$ into the latent representations $c_t^i$, 
while using the proprioception encoder $e_p$ with parameter $\theta_p$ to extract the representation $c_t^p$ from the proprioceptive observation $o_t^p$. We then concatenate the representations of the image and proprioception and use a fusion model $f_\upsilon$  parametrized by a multilayer perceptron (MLP) with parameter $\theta_\upsilon$ to transfer the concatenated representation into the joint representation $j_t$.  A stochastic
neural network with parameter $\theta_\psi$ subsequently is used to transfer the joint representation $j_t$ to the mean and standard error of the diagonal Gaussian distribution, which is used to randomly sample the bottleneck variable $z_t$. We denote this Gaussian distribution as $d_\psi(z_t|c_t^p, c_t^i)$, since the generated variable $z_t$ is conditioned on the image and proprioception representations. The prediction head $m_\eta$ with parameters $\theta_\eta$ and the projection head $l_\delta$ with parameters $\theta_\delta$ both induce nonlinear transformation on $z_t$ and $a_t$, which is used to estimate the lower bound on $I(z_t, a_t; z_{t+1})$. When generating the next image representation $c_{t+1}^i$,  state representation $c_{t+1}^p$ and bottleneck variable $z_{t+1}$, we use target networks for stopping gradients through the
image and proprioception encoders, the fusion model, and the stochastic encoder. The parameters of target networks are updated by using the exponential moving average (EMA) of their corresponding online networks, which has been proven to be effective in improving the stability of training~\citep{he2020momentum}.  

\subsection{Minimizing An Upper Bound of the MIB Objective}
In this section, we derive an upper bound of our MIB objective for tractable optimization. We firstly obtain an upper bound of $I(z_t;c_t^p, c_t^i)$ by using the variational distribution $q(z_t)$ to approximate the marginal distribution $p(z_t)$ in the definition of $I(z_t;c_t^p, c_t^i)$, 
\begin{equation}
\begin{aligned}
I(z_t;c_t^p, c_t^i) &= \mathbb{E}_{p(z_t,c_t^p, c_t^i)} \bigg[\log \frac{p(z_t|c_t^p, c_t^i)}{p(z_t)}\bigg]\\
&= \mathbb{E}_{p(z_t,c_t^p, c_t^i)} \bigg[\log \frac{p(z_t|c_t^p, c_t^i)}{q(z_t)}\bigg] - \mathbb{D}\textsubscript{KL} \big( p(z_t) \parallel q_\mu(z_t) \big)\\
& \leq \mathbb{E}_{p(z_t,c_t^p, c_t^i)} \bigg[\log \frac{p(z_t|c_t^p, c_t^i)}{q(z_t)}\bigg]\\
&= \mathbb{E}_{p(c_t^p, c_t^i)} \bigg[\mathbb{D}\textsubscript{KL} \big( d_\psi(z_t|c_t^p, c_t^i) \parallel q(z_t) \big)\bigg] = I_\text{upper}\rlap{,}
\end{aligned}
\label{eq:up bound}
\end{equation}
where the inequality is introduced by the non-negativity of KL divergence. In practice, we use a unit normal Gaussian distribution $q_(z_t) = \mathcal{N}(0, \mathbf{I})$ as the variational distribution. The obtained upper bound, which is the expected KL divergence between two diagonal Gaussian distributions, can be computed in a closed form.

To effectively estimate the predictive information $I(z_t, a_t; z_{t+1})$, we employ the InfoNCE lower bound on the $I(z_t, a_t; z_{t+1})$~\citep{oord2018}, which is constructed as follows, 
\begin{equation}
\begin{aligned}
I(z_t, a_t; z_{t+1}) &\geq  \underset{p(z_t, a_t, z_{t+1}), N}{\mathbb{E}}  \bigg[\log\frac{\exp\big(h_\omega(z_t, a_t, z_{t+1})\big)}{\sum_{z_{t+1}^* \in N \cup z_{t+1}}\exp\big(h_\omega(z_{t+1}^*, z_{t}, a_t)\big)}\bigg] \\
& = I_\text{lower}\rlap{,}
\end{aligned}
\label{lb}
\end{equation}
where the positive sample pair $(z_t, a_t, z_{t+1})$ are drawn from the joint distribution $p(z_t, a_t, z_{t+1})$ and $N$ denotes a set of negative samples $z_{t+1}^*$ sampled from the marginal distributions $p(z_{t+1})$. The score function $h_\omega(\cdot, \cdot)$ with parameters $\theta_\omega$ that aims to discriminate the positive sample pairs from the negative sample pairs $(z_{t+1}^*, a_t, z_{t+1})$,  is given by 
\begin{equation}
\begin{aligned}
h_\omega(z_t, a_t, z_{t+1})= l_\delta\big(m_\eta(z_t, a_t)\big)^\top  \omega l_\delta^-(z_{t+1}),
\end{aligned}
\label{score function}
\end{equation}
where $\omega$ is the learnable transformation matrix, and the projection head $l_\delta(\cdot)$ and the prediction head $m_\eta(z_t, a_t)$ nonlinearly transform the concatenated $z_t$ and $a_t$ into the latent space where the contrastive estimation is applied. While mapping $z_{t+1}$ into the same latent space, we use the target projection head $l_\delta^-(\cdot)$, whose parameters are updated by the exponential moving average of the projection head $l_\delta(\cdot)$. 

By combining the upper bound in Eq.~\ref{eq:up bound} and the lower bound in Eq.~\ref{lb}, we can obtain an upper bound of our MIB objective for tractable optimization, 
\begin{equation}
\begin{aligned}
\underset{\theta_i, \theta_p, \theta_\upsilon, \theta_\psi, \omega, \theta_\delta, \theta_\eta}{\argmin} \quad \mathcal{L} &= \alpha \underset{p(c_t^p, c_t^i)}{\mathbb{E}} \bigg[\mathbb{D}\textsubscript{KL} \big( d_\psi(z_t|c_t^p, c_t^i) \parallel q(z_t) \big) \bigg] \\
& \quad -  \underset{p(z_t, a_t, z_{t+1}), N}{\mathbb{E}}  \bigg[\log\frac{\exp\big(h_\omega(z_t, a_t, z_{t+1})\big)}{\sum_{z_{t+1}^* \in N \cup z_{t+1}}\exp\big(h_\omega(z_{t+1}^*, z_{t}, a_t)\big)}\bigg]. 
\end{aligned}
\label{loss}
\end{equation}
During training, we approximate the expectations in the upper bound via drawing samples from the replay buffer. We jointly optimize the parameters of the image encoder~$\theta_i$, the proprioception encoder~$\theta_p$, the fusion model~$\theta_\upsilon$, the stochastic encoder~$\theta_\psi$, the matrix~$\omega$, the projection head $\theta_\delta$, and the prediction head $\theta_\eta$ by minimizing the upper bound using stochastic gradient descent.

\subsection{Policy and Value Function Learning}

For learning the policy and value function, we adopt the SAC~\citep{haarnoja2018soft} with the only difference that the policy and the value function use the joint representation $j_t$ as the input rather than the raw images and proprioceptive data. We use the intermediate joint representations  $j_t$ because they empirically perform better than their corresponding bottleneck representations, which is consistent with prior work~\citep{lee2020predictive}. We employ the SAC algorithm to learn optimal policies, since offline actor-critic algorithms have been successful in solving complex optimal control tasks~\citep{wang2023recent, wang2023adaptive}. During training, we allow the gradient of the value function through the stochastic encoder, the fusion model, and observation encoders by following~\citep{yarats2019improving}.




Algorithm~\ref{alg:MIB} shows the training algorithm of our method. The training algorithm proceeds by alternating between collecting experience by interacting with the environment, and optimizing the SAC agent and the upper bound of the MIB objective. We optimize the parameters of the policy and value function networks by the actor loss and critic loss proposed by~\citep{haarnoja2018soft}. The parameters of our MIB model are optimized by Eq.\ref{loss}. 

\begin{algorithm*}[ht]
\SetAlgoLined
\textbf{Given:} learnable parameters $\theta_i$, $\theta_p$, $\theta_\upsilon$, $\theta_\psi$, $\omega$, $\theta_\delta$, $\theta_\eta$, replay buffer $\mathcal{D}$, batch size $B$\\
 Initialize replay buffer $\mathcal{D}$\\
 \For{each training step}{
  Collect transition $(s_t, a_t, r_t, s_{t+1})$ and add it to the replay buffer $\mathcal{D}$\\
  \For{each gradient step}{
   Sample a minibatch of transitions: $\{s_t, a_t, s_{t+1}, r_t\}_1^B \sim \mathcal{D}$.\\
   Generate joint representations and bottleneck representations by Eq.\ref{eq:encoding}\\
   Update actor and critic networks of SAC\\
   Update observation encoders, fusion model, stochastic encoder, transformation matrix, projection head, prediction head by Eq.\ref{loss}\\
   Update target encoders by EMA
  }
 }
 \caption{Training Algorithm for MIB}
 \label{alg:MIB}
\end{algorithm*}

\section{Experimental Evaluation}
\label{exp}

\subsection{Experimental Setup}

\begin{figure*}
    \centering
    \makebox[\textwidth]{\includegraphics[width=80mm]{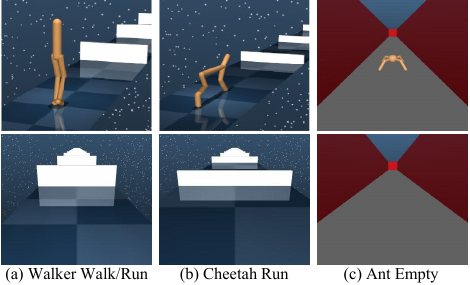}}
    \caption{Continuous locomotion tasks are used in our experiments, namely Hurdle Walker Walk/Run, Hurdle Cheetah Run, and Ant Empty. Three-person views for visualization and egocentric images provided to the agent are shown in the upper and bottom rows, respectively.}
    \label{fig:dmc tasks}
\end{figure*}

We evaluate our MIB method on several modified challenging locomotion tasks from the DMC~\citep{tassa2018deepmind} following~\citep{becker2023reinforcement}. For each task, the agent needs to integrate egocentric images with proprioception for control. Specifically, we evaluate our method on four continuous locomotion control tasks, namely Hurdle Walker Walk, Hurdle Walker Run, Hurdle Cheetah Run, and Ant Empty (Fig.~\ref{fig:dmc tasks}). The first three tasks are achieved by adding hurdles to standard Walker Walk, Walker Run, and Cheetah Run tasks from the DMC, while the last one is readily available in the DMC. In Hurdle Walker Walk, Hurdle Walker Run, and Hurdle Cheetah Run tasks, the goal of the agent is to control a robot with complex dynamics and contacts with the ground to step over hurdles to move forward. In the Ant Empty task, the agent needs to control a robot to run through the corridor as fast as possible. Egocentric images provide the agent with details of obstacles (hurdles or corridors) so that the robot can avoid them, while proprioception allows the agent to be aware of the internal states of the robot.   

We further evaluate the zero-shot robustness of the joint representations and policies learned by our method. The aim of this experiment is to investigate whether the latent joint representations and policies learned by our method are robust to unseen observation perturbations, such as noise or changes in backgrounds. We carry out the evaluation by firstly learning the representations and policies in locomotion tasks without distractive information, and then testing the learned models in tasks with noises and natural backgrounds. In tasks with natural backgrounds, we corrupt the egocentric images in the locomotion tasks with natural videos sampled from the Kinetics dataset by following~\citep{chen2021multi}. For tasks with noises, we add Gaussian white noise to proprioception.  Gaussian noises and natural backgrounds both introduce unseen and task-irrelevant patterns to raw observations.


We compare MIB to the following leading baselines in our experiments: MuMMI~\citep{chen2021multi}, which extracts latent joint representations from multimodal observations by maximizing the mutual information between joint representations and representations of single modalities, DB~\citep{bai2021dynamic}, which uses variational information bottleneck to extract compressed representations solely from images, and Drq~\citep{yarats2021image} which learns image representations by improving the robustness of value function to image augmentation and achieves promising performance on control tasks from DMC. To ensure a fair comparison, we concatenate the embeddings of the proprioception with representations trained solely on images by DB and Drq, respectively. Moreover, we also compare our algorithm with Vanilla SAC~\citep{haarnoja2018soft},  which directly operates from raw egocentric images and proprioception without an auxiliary loss of representation learning. We provide the implementation details of our method and baselines in the Appendix.


\begin{figure*}
    \centering
    \makebox[\textwidth]{\includegraphics[width=\textwidth]{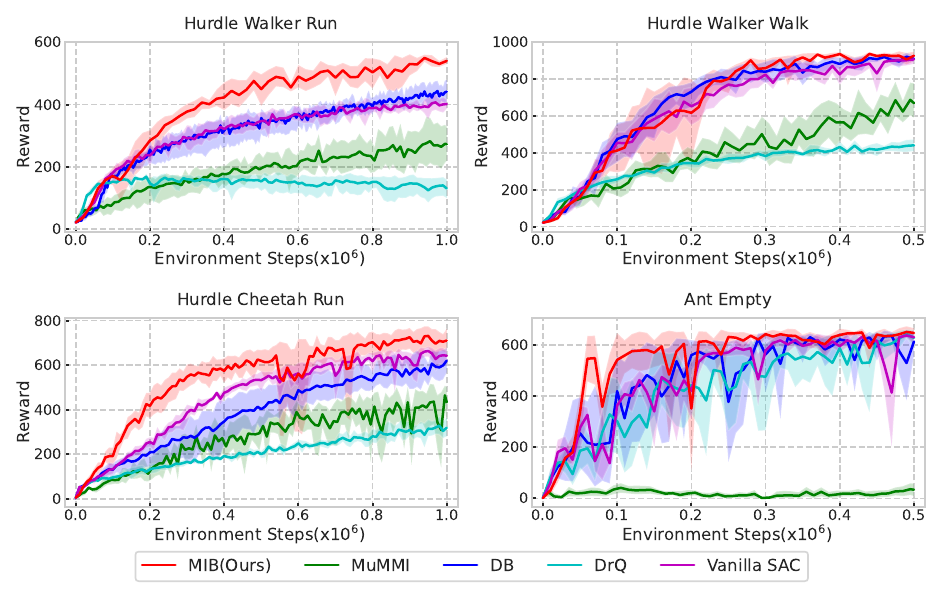}}
    \caption{We compare the performance of our method to baselines on four challenging locomotion tasks. The plot shows the average reward and 95\% confidence interval across 5 independent runs. For each run, the mean return is computed over 10 trajectories. Our method achieves better sample efficiency and performance than all baselines across all tasks. }
    \label{fig:sample efficiency}
\end{figure*}

\begin{table*}[b!]
\caption{Rewards achieved by our method and baselines (mean and standard error for 5 seeds) at 500K environment steps on Hurdle Walker Walk and Ant Empty that are relatively simple tasks, and 1000K environment steps on Hurdle Walker Run and Hurdle Cheetah Run. Bold font means the highest average reward among these methods. Our method achieves higher average rewards than baselines.}
\begin{center}
\sisetup{%
            table-align-uncertainty=true,
            separate-uncertainty=true,
            detect-weight=true,
            detect-inline-weight=math
        }
\resizebox{\textwidth}{!}{\begin{tabular}{c | c c c c c c | c}
    \Xhline{2\arrayrulewidth}
    Scores & MIB(Ours) & DB & MuMMI & DrQ & Vanilla SAC\\
    \hline
    Hurdle Walker Run & \bfseries 540$\pm$ \bfseries 6  &  442 $\pm$ 18  & 273$\pm$29 & 130$\pm$17 & 401$\pm$8\\
    Hurdle Walker Walk & \bfseries 928$\pm$ \bfseries 13 & 911 $\pm$  9  & 671$\pm$50 & 441$\pm$ 16 & 908$\pm$ 11\\
    Hurdle Cheetah Run & \bfseries 711$\pm$ \bfseries28  & 620 $\pm$ 30  & 443$\pm$45 & 319$\pm$ 22 & 642$\pm$ 9\\
    Ant Empty & \bfseries 648$\pm$ \bfseries 12  & 614 $\pm$ 23 & 34$\pm$12 & 629$\pm$ 18 & 631$\pm$ 15\\
    \Xhline{2\arrayrulewidth}
\end{tabular}}
\end{center}
\label{table:performance}
\end{table*}

\subsection{Sample Efficiency and Performance}
Fig~\ref{fig:sample efficiency} and Table~\ref{table:performance} compare our algorithm with MuMMI, DB, Drq, and  Vanilla SAC on four challenging locomotion tasks with egocentric images and proprioception. Our algorithm outperforms all baselines on all tested tasks in terms of sample efficiency. Moreover, on Hurdle Walker Run and Hurdle Cheetah Run tasks,  MIB outperforms baselines by a large margin. Notably, MIB outperforms Vanilla SAC on all tasks. For example, MIB achieves a return of 711, while Vanilla SAC achieves a return of 642 on the Cheetah Run task. The performance gain achieved by our method compared to  Vanilla SAC verifies that the learned joint representations by our MIB objective help learn good policies on challenging locomotion tasks.

\subsection{Zero-shot Robustness}
We further investigate whether the learned representations and policies learned by our method and baselines are robust to unseen observation perturbations without additional fine-tuning. In the presence of white noises and natural backgrounds that are unseen during training, we test the performance of the representations and policies learned by our method and baselines at a fixed number of environmental steps (500K for relatively simple Hurdle Walker Walk and Ant Empty tasks and 1000K for Hurdle Walker Walk and Hurdle Cheetah Run tasks). We compare the zero-shot robustness of MIB against MuMMI, DB, Drq, and  Vanilla SAC on locomotion tasks with unseen white noise in the top half of Table~\ref{table:noisy envs}. MIB outperforms all competing approaches by a large margin on the majority of tasks.  For example,  MIB achieves a return of 642, which is better than 590 achieved by DB  and 577 achieved by Drq on Ant Empty with noise. The results on locomotion tasks with noise indicate that MIB is robust in the presence of white noise.  In the bottom half of Table~\ref{table:noisy envs}, we compare the zero-shot robustness to natural backgrounds. Although MIB degrades in the presence of unseen natural backgrounds, it remains more robust than baselines, except for Hurdle Cheetah Run.

\begin{table*}[t!]
\centering
\caption{Scores achieved by our method and baselines (mean and standard error for 5 seeds) on four locomotion tasks with white noise or natural backgrounds that are unseen before. Bold font means the highest average reward among these methods. Our method performs best on the majority of tasks.}
\label{table:noisy envs}
\sisetup{%
            separate-uncertainty=true
        }
\renewrobustcmd{\bfseries}{\fontseries{b}\selectfont}
\renewrobustcmd{\boldmath}{}
\resizebox{\textwidth}{!}{
\begin{tabular}{c| c | c c c c c}
    \Xhline{2\arrayrulewidth}
    \multicolumn{2}{c|}{Scores} & MIB(Ours) & DB & MuMMI & DrQ & Vanilla SAC\\
    \hline
    \multirow{4}{*}{White Noise} & Hurdle Walker Run & \bfseries 531$\pm$ \bfseries 6  &  438 $\pm$  6  & 162$\pm$26 & 128$\pm$16 &403$\pm$ 7\\
    & Hurdle Walker Walk & \bfseries 940$\pm$ \bfseries 9 & 912 $\pm$  6  & 346$\pm$36 & 435$\pm$ 10 &870$\pm$ 30\\
    & Hurdle Cheetah Run & \bfseries 656$\pm$ \bfseries 31  & 592 $\pm$ 8  & 65$\pm$18 & 309$\pm$ 14 &634$\pm$ 10\\
    & Ant Empty & \bfseries 642$\pm$ \bfseries 9  & 590 $\pm$ 12  & 23$\pm$9 & 577$\pm$ 28 &609$\pm$ 18\\
    \hline
    \multirow{4}{*}{Natural background}  & Hurdle Walker Run & \bfseries 312$\pm$ \bfseries 23  &  258 $\pm$  22 & 171$\pm$ 25 & 103$\pm$8 & 217$\pm$ 13\\
    & Hurdle Walker Walk & \bfseries 783$\pm$ \bfseries 24  & 645 $\pm$  42 & 346$\pm$31 & 247$\pm$ 19 & 694$\pm$ 50\\
    & Hurdle Cheetah Run & 101$\pm$  11  & \bfseries 129 $\pm$ \bfseries 8 & 81$\pm$11 & 82$\pm$ 6 & 96$\pm$ 7\\
    & Ant Empty & \bfseries 563$\pm$ \bfseries 24  & 508 $\pm$ 9 & 18$\pm$8 & 58$\pm$ 3 & 523$\pm$ 40\\
    \Xhline{2\arrayrulewidth}
\end{tabular}}
\end{table*}

\begin{table*}[t!]
\centering
\caption{Scores achieved by our method and its ablations (mean and standard error for 5 seeds) on four locomotion tasks w/o unseen white noise. Bold font means the highest average reward among these methods. Our method outperforms all ablations on all tasks in terms of average reward.}
\label{table:ablation}
\sisetup{%
            separate-uncertainty=true
        }
\renewrobustcmd{\bfseries}{\fontseries{b}\selectfont}
\renewrobustcmd{\boldmath}{}
\resizebox{\textwidth}{!}{
\begin{tabular}{c| c | c c c c c}
    \Xhline{2\arrayrulewidth}
    \multicolumn{2}{c|}{Scores} & MIB(Ours) & Non-KL & Non-Loss & Non-Img & Non-Prop\\
    \hline
    \multirow{4}{*}{Without Noise} & Hurdle Walker Run & \bfseries 540$\pm$ \bfseries 6  &  509 $\pm$  19  & 331$\pm$14 & 396$\pm$14 &49$\pm$ 9\\
    & Hurdle Walker Walk & \bfseries 928$\pm$ \bfseries 13 & 911 $\pm$  17  & 744$\pm$20 & 912$\pm$ 5 &152$\pm$ 11\\
    & Hurdle Cheetah Run & \bfseries 711$\pm$ \bfseries 28  & 562 $\pm$ 34  & 385$\pm$19 & 288$\pm$ 30&119$\pm$ 7\\
    & Ant Empty & \bfseries 648$\pm$ \bfseries 12  & 640 $\pm$ 9  & 98$\pm$59 & 221$\pm$ 35 &116$\pm$ 25\\
    \hline
    \multirow{4}{*}{With Noise}  & Hurdle Walker Run & \bfseries 531$\pm$ \bfseries 6  &  505 $\pm$  18 & 337$\pm$11 &  396 $\pm$  13 & 31$\pm$3 \\
    & Hurdle Walker Walk & \bfseries 940$\pm$ \bfseries 9 & 906 $\pm$  27 & 704$\pm$ 35 & 899 $\pm$  12 & 42$\pm$5\\
    & Hurdle Cheetah Run & \bfseries 656$\pm$ \bfseries 31  & 539 $\pm$ 23 & 387$\pm$13 & 282 $\pm$ 21 & 9$\pm$1\\
    & Ant Empty & \bfseries 642$\pm$ \bfseries 9  & 637 $\pm$ 5 & 80$\pm$ 53 & 223 $\pm$ 13  & 91$\pm$15\\
    \Xhline{2\arrayrulewidth}
\end{tabular}}
\end{table*}

\subsection{Empirical Analysis}

In this section, we empirically analyze why our method works well. We perform ablation studies to investigate the individual contributions of the mutual information $I(z_t;c_t^p, c_t^i)$ and the predictive information $I(z_t, a_t; z_{t+1})$. We also perform several experiments to study whether utilizing information from multiple sensors for learning representations and policies improves the performance compared to using only a single sensor modality. Specifically, we investigate four ablations of our MIB model: Non-KL, which removes the first term in Eq.~\ref{loss} that is an upper bound on $I(z_t;c_t^p, c_t^i)$, Non-Loss, which removes the total objective in Eq.~\ref{loss} including an upper bound of $I(z_t;c_t^p, c_t^i)$ and a lower bound of $I(z_t, a_t; z_{t+1})$, Non-Img, which uses representations solely extracted from proprioception, and Non-Prop, which uses embeddings only extracted from egocentric images.

Table~\ref{table:ablation} compares the performance and zero-shot robustness of MIB and its four ablations on four locomotion tasks with or without white noise that they have not seen before. The results show that MIB achieves better performance and zero-shot robustness to noise than four ablations on all tasks. Specifically, MIB outperforms Non-KL on all tasks, which indicates that compressing the raw multimodal observations improves the performance and zero-shot robustness to noise. By comparing Non-KL and No-Loss, we can conclude that preserving the predictive information by maximizing the lower bound on $I(z_t, a_t; z_{t+1})$ can significantly improve the performance. The performance gain achieved by our method compared to Non-Img and Non-prop indicates that our joint representations exploiting information from multiple sensors can achieve better performance on locomotion tasks than representations extracted solely from a single sensory input.

\section{Conclusion and Future work}
\label{conclusion}
We presented an information-theoretic multimodal representation learning approach, MIB, which learns compressed joint representations from egocentric images and proprioception for improving the sample efficiency and performance of reinforcement learning agents. The objective of our MIB model is to compress irrelevant information in raw multimodal observations and meanwhile preserve the complementary and relevant information for learning a policy. Empirical evaluations on a set of challenging locomotion tasks with egocentric images and proprioception show that our algorithm achieves better sample efficiency than leading methods. Furthermore, we observed that the learned representations and policies by our method achieve better zero-shot robustness to unseen white noise and natural backgrounds compared to other methods, and using multimodal observations is more useful for learning locomotion policies than using individual modalities.

Although our algorithm is comparable to leading baselines with auxiliary losses of representation learning in terms of wall-clock time, it slightly increases time cost than Vanilla SAC. In addition, our method relies on an iterative optimization process and therefore has no guarantee to find the optimal joint representations that contain the minimal and sufficient information for future prediction. In future work, we will extend our MIB model into tasks with different sensory inputs, such as tactile and audio feedback. Integrating more powerful transformer architectures to extract joint representations from time-series multimodal observations is also a promising direction for our future research.

\section*{Data availability}
The code and data will be made available on request.

\section*{Acknowledgement}

This work was supported by the National Natural Science Foundation of China under Grants 62025304.

\bibliography{reference}

\newpage
\appendix
\setcounter{table}{0}
\renewcommand\thetable{\Alph{section}.\arabic{table}}
\section{Implementation Details}
\subsection{Network Architecture} 
In this section, we present the network architecture of our MIB model as follows.
\begin{itemize}
    \item \textbf{Image encoder} We use the convolutional encoder architecture from SAC-AE~\citep{yarats2019improving} to parameterize the image encoder. The image encoder $e_{i}$ consists of four convolutional layers with 32 channels and $3\times3$ kernels and a stride of 1. Its output dimension is set to 50.
    \item \textbf{Proprioception encoder} The proprioception encoder $e_{p}$ is parameterized by a 2-layer fully-connected network with FCN(units=512) $\rightarrow$ FCN(units=50) architecture and ReLU hidden activations. 
    \item \textbf{Fusion model} The fusion model $f_\upsilon$ consists of two fully-connected layers with FCN(units=1024) $\rightarrow$ FCN(units=50) architecture and ReLU hidden activations. 
    \item \textbf{Stochastic encoder} The stochastic encoder consists of two fully-connected layers with FCN(units=1024) $\rightarrow$ FCN(units=100) architecture and ReLU hidden activations. Its output is divided into the mean $\mu \in \mathbb{R}^{50}$ and the standard deviation $\sigma \in \mathbb{R}^{50}$ of the diagonal Gaussian distribution $d_\psi(z_t|c_t^p, c_t^i)$.
    \item \textbf{Prediction head} The prediction head $m_\eta$ is just a linear layer which maps the concatenated representation $z_t \in \mathbb{R}^{50}$ and action $a_t \in \mathbb{R}^{|\mathcal{A}|}$ into a 50-dimensional vector.
    \item \textbf{Projection head}  The projection head $l_\delta$ is parameterized by a 2-layer fully-connected network with FCN(units=1024) $\rightarrow$ FCN(units=50) architecture and ReLU hidden activations. 
\end{itemize}

The target networks share the same network architectures with their corresponding online networks.

\subsection{Image Preprocessing}
We construct an individual image observation $o_t^{i}$ by stacking 3 consecutive frames, where each frame is an egocentric image with size $84 \times 84 \times 3$ rendering from the 2nd camera.  We then divide each pixel value by 255 and scale it down to  $[0, 1]$ range. Before we feed images into the encoders, we follow~\citep{yarats2021image} by performing data augmentation by randomly shifting the egocentric image by $[-4, 4]$.

\begin{table}[t!]
    \centering
    \caption{Hyperparameters used in MIB}
    \label{table: SAC hyper}
    \begin{tabular}{l l}
        \Xhline{2\arrayrulewidth}
        Parameter & Value\\
        \hline
        Replay buffer capacity & 100 000\\
        Optimizer & Adam\\
        Critic Learning rate & $10^{-3}$ \\
        Critic  Q-function EMA & 0.01\\
        Critic target update freq & 2\\
        Actor learning rate & $10^{-3}$ \\
        Actor update frequency & 2\\
        Actor log stddev bounds & [-10 2]\\
        Temperature learning rate & $10^{-3}$ \\
        Initial steps & 1000\\
        Discount & 0.99\\
        Initial temperature & 0.1\\
        Batch Size & 128 (Hurdle Walker) \\
         & 512 (Hurdle Cheetah, Ant)\\
        Learning rate for MIB & $10^{-4}$\\
        Encoder EMA $\tau$ & 0.05 \\
        Coefficient $\alpha$ & 0.0001\\
        Action repeat  & 2 (Hurdle Walker, Ant) \\
         & 4 (Hurdle Cheetah) \\
        \Xhline{2\arrayrulewidth}
    \end{tabular}
\end{table}

\subsection{SAC Implementation}
We use the common implementation of SAC introduced by~\citep{yarats2019improving}.  We use all hyperparameters of SAC from \citep{yarats2019improving}, except for the replay buffer capacity, which we set to a smaller $10^5$  rather than $10^6$. Following~\citep{you2022integrating}, we set the learning rate for the gradient update of MIB and the coefficient of the exponential moving average to $10^{-4}$ and 0.05, respectively. We treat batch size and the number of action repeat as hyperparameters of the agent by following common practice~\citep{you2022integrating, hafner2019learning}. The only specific hyperparameter of MIB, namely the coefficient $\alpha$, was determined by performing hyperparameter tuning on the Hurdle Walker Run task and then fixed across all tasks. We show all used hyperparameters in Table~\ref{table: SAC hyper}.

\subsection{Baseline Implementation}
We next describe the implementation of baselines and the modifications we made to ensure a fair comparison in our experiments.
\begin{itemize}
    \item \textbf{DrQ} We obtain the results for Drq by performing the official implementation provided by~\citep{yarats2019improving} with one difference. We concatenate image embeddings with proprioceptive representations and feed the concatenated embeddings to the actor and the policy to make a fair comparison to our setup. The encoder for extracting representations from proprioception shares the same network architecture with our proprioception encoder and is optimized by the critic's gradient. 
    \item \textbf{DB} We adapt DB to the continuous control settings with egocentric images and proprioception. To facilitate this, we augmented SAC~\citep{yarats2019improving} with DB model~\citep{bai2021dynamic} that takes the concatenation of image representations and proprioceptive representations as inputs. Similar to MIB, we employ the convolutional encoder from~\citep{yarats2019improving} to encode images and the proprioception encoder parameterized by a 2-layer MLPs with FCN(units=512) $\rightarrow$ FCN(units=50) architecture with ReLU hidden activations to encode proprioception for DB.
    \item \textbf{Vanilla SAC} We implement Vanilla SAC by performing the common implementation of SAC~\citep{yarats2019improving} with one difference that the agent inputs the concatenation of image representations and proprioceptive representations. Without additional loss, the image and proprioception encoders are optimized purely by the critic's gradient.
    \item \textbf{MuMMI} We use the official implementation provided by~\citep{chen2021multi} to obtain the results for MuMMI. The agent combines the representations of images and proprioception to make decisions.
\end{itemize}

\end{document}